\documentclass{article}

\PassOptionsToPackage{numbers, compress}{natbib}

\usepackage[preprint,nonatbib]{neurips_2023}

\usepackage{longtable}
\usepackage[utf8]{inputenc} 
\usepackage[T1]{fontenc}   
\usepackage{hyperref}      
\usepackage{url}            
\usepackage{booktabs}       
\usepackage{amsfonts}       
\usepackage{CJKutf8}
\usepackage{nicefrac}       
\usepackage{array}
\usepackage{microtype}      
\usepackage{xcolor}        
\usepackage{amsmath}
\usepackage{tcolorbox}
\usepackage{makecell} 
\usepackage{array}    
\usepackage{float}
\usepackage{graphicx}
\usepackage{ltablex}
\usepackage{lipsum}
\usepackage{graphicx}
\usepackage{wrapfig}
\usepackage[size=small]{caption}
\usepackage{mathtools}
\usepackage{tabularx}
\usepackage{amssymb}
\usepackage{amsthm}
\usepackage{soul}
\usepackage{float}

\definecolor{textblue}{rgb}{.2,.2,.7}
\definecolor{textred}{rgb}{0.54,0,0}
\definecolor{textblack}{rgb}{0,0,0}
\definecolor{textgreen}{rgb}{0,0.53,0}
\usepackage{listings}
\lstdefinestyle{pythonstyle}{
    language=Python,
    morekeywords={None}, 
    breaklines=true,
}

\usepackage{array}
\newcolumntype{M}[1]{>{\centering\arraybackslash}m{#1}}

\title{Squid: Long Context as a New Modality for Energy-Efficient On-Device Language Models}

\author{%
  Wei Chen\\
  Nexa AI\\
  Sunnyvale, CA 94086 \\
  \texttt{alexchen@nexa4ai.com} \\
  \And
  Zhiyuan Li\\
  Nexa AI\\
  Sunnyvale, CA 94086 \\
  \texttt{zack@nexa4ai.com}
  \AND
  \hspace{0.5cm}Shuo Xin \\
  \hspace{0.5cm}Nexa AI\\
  \hspace{0.5cm}Sunnyvale, CA 94086 \\
  \hspace{0.5cm}\texttt{shuo@nexa4ai.com} \\
  \And
  Yihao Wang \\
  Nexa AI \\
  Sunnyvale, CA 94086 \\
  \texttt{ethan@nexa4ai.com} \\
}

\begin{document}
\begin{CJK*}{UTF8}{gbsn}
\nocite{*}
\maketitle

\begin{abstract}
This paper presents Squid, a novel decoder-decoder architecture for energy-efficient processing of long contexts in language models. Our approach addresses the significant energy consumption and latency challenges inherent in on-device models. 
Squid employs a compact 0.5B-parameter decoder to distill extensive contextual information into a memory embedding, substantially reducing the input length for the primary 7B-parameter decoder model. Inspired by vision-language models, we repurpose the image embedding projector to encode long textual contexts, effectively treating extended context as a distinct modality. This innovative method enables the processing of substantially longer contexts without the typical computational overhead associated with extended input sequences. 
Empirical evaluations demonstrate a 10-fold improvement in energy efficiency and a 5-fold reduction in latency compared with conventional full-length context processing, without any loss in response quality. Our work contributes to the development of more sustainable and scalable language models for on-device applications, addressing the critical need for energy-efficient and responsive AI technologies in resource-constrained environments while maintaining the accuracy required to understand long contexts. 
This research has implications for the broader field of natural language processing, particularly in the domain of efficient model design for resource-limited settings. By enabling more sophisticated AI capabilities on edge devices, Squid paves the way for advanced language processing in a wide range of applications where computational resources are at a premium. The Squid model is publicly available at \href{https://huggingface.co/NexaAIDev/Squid}{https://huggingface.co/NexaAIDev/Squid}.

\end{abstract}

\section{Introduction}
On-device language models have become increasingly crucial in our interconnected world, offering enhanced privacy, reduced latency, and offline functionality~\cite{gerganov_llama_cpp, mlc-llm, lugaresi2019mediapipe, pytorch2023executorch, OpenLLM, song2023powerinfer, chen2024octopusv2ondevicelanguage}. However, these models face significant challenges, particularly in energy consumption and processing speed when handling long contexts. Battery life on mobile devices is a critical concern, as complex language processing tasks can rapidly deplete power resources, limiting the practical utility of on-device AI applications. This energy constraint is further exacerbated when processing long contexts, which require more computational resources and memory. Moreover, the latency introduced by processing extensive input sequences can severely impact user experience, especially in real-time applications such as voice assistants or interactive chatbots. Consequently, there is an urgent need for innovative approaches that maintain the accuracy and capability of language models while significantly reducing their energy footprint and improving response times.

To address these challenges, various approaches have been developed to mitigate the context length problem in large language models (LLMs). Retrieval-Augmented Generation (RAG)~\cite{NEURIPS2020_6b493230} has emerged as a prominent solution, incorporating an external retrieval component to search for relevant information, thereby allowing the model to handle extensive knowledge without storing all information in its parameters. Recent advancements, such as the LongRAG~\cite{jiang2024longrag} framework, have further improved this approach by balancing the workload between the retriever and the reader, enabling the processing of much larger token inputs. Another effective strategy focuses on optimizing the key-value (KV) cache. Techniques such as chunk-wise KV cache compression and swapping have been implemented to minimize context-switching overhead and enable efficient state maintenance across multiple invocations. The LLMaaS (Language Models as a Service)~\cite{yin2024llmservicemobiledevices} paradigm exemplifies this approach by integrating LLMs as system services on mobile devices, employing stateful execution to maintain persistent states and reduce memory usage. While these methods have shown promise, they often involve trade-offs between context length, model performance, and computational efficiency, highlighting the need for more holistic solutions.

Other works have sought to directly reduce the length of the context to lower computational costs~\cite{chevalier2023adapting, NEURIPS2023_3d77c6dc, ge2024incontext}. Although these approaches aim to reduce computational costs via context compression, the compression step itself can still introduce overhead, and they do not address the alignment issue between the compressed context and the original text.

In response to these challenges, we introduce Squid, a novel decoder-decoder architecture designed specifically for energy-efficient processing of long contexts in language models. We were inspired by recent work on Vision-Language Models (VLMs)~\cite{NEURIPS2023_6dcf277e, Liu_2024_CVPR, Lin_2024_CVPR, bai2023qwenvlversatilevisionlanguagemodel, lu2024deepseekvlrealworldvisionlanguageunderstanding}, which demonstrates that model performance can benefit from specially designed multi-stage training procedures. Our approach utilizes a small 0.5B decoder to distill extensive contextual information into several memory tokens, significantly reducing the input length for the primary 7B decoder model. Drawing inspiration from vision-language models, we repurpose the image embedding projector to encode long textual contexts, effectively treating extended context as a distinct modality. This innovative method enables the processing of substantially longer contexts without incurring the typical computational overhead associated with extended input sequences. By doing so, Squid achieves an impressive 10-fold improvement in energy efficiency and a 5-fold reduction in latency compared with conventional full-length context processing methods. Our work contributes to the development of more sustainable and scalable language models for on-device applications, addressing the critical need for energy-efficient and responsive AI technologies in resource-constrained environments while maintaining the accuracy required to understand long contexts. This breakthrough has far-reaching implications for the deployment of sophisticated AI capabilities on edge devices, potentially revolutionizing fields such as mobile computing, IoT, and wearable technology.

\section{Related Work}

\textbf{Prompt compression}\quad Prompt compression has emerged as a critical area of research to address the challenges associated with long-context inference in LLMs. Existing methods can be divided into three main categories: token pruning, abstractive compression, and extractive compression. Token pruning techniques, such as LongLLMLingua~\cite{jiang2023longllmlingua} and Selective-Context~\cite{li2023compressing}, aim to reduce prompt length by removing less important tokens. Abstractive compression methods, including RECOMP~\cite{xu2023recomp} and Prompt-SAW~\cite{ali2024prompt}, utilize summarization techniques to condense the original context. Extractive compression methods, such as RECOMP's extractive component and document rerankers~\cite{nogueira2019passage, pradeep2023rankvicuna, pradeep2023rankzephyr}, select relevant documents, sentences, or phrases from the original context. These approaches can be further classified as query-aware or query-agnostic, depending on whether they tailor compression based on the specific question or task. 
Other works directly compress or distill the context to lower computational costs. AutoCompressor~\cite{chevalier2023adapting} recursively compresses long contexts into compact summary vectors. CEPE~\cite{yen2024long} introduced parallel encoding to extend the context window. Tan et al. address long contexts through offline learning in LLoCO~\cite{tan2024lloco}. StreamingLLM~\cite{xiao2024efficient} and Unlimiformer~\cite{bertsch2023unlimiformer} modify the attention mechanism to achieve a longer context window. REPLUG~\cite{shi-etal-2024-replug} uses a retrieval model to process contexts separately. Mu et al. proposed gist tokens~\cite{NEURIPS2023_3d77c6dc}, which compress context length through context distillation and modifications to the attention mask. ICAE~\cite{ge2024incontext} uses an encoder, fine-tuned from an LLM via LoRA~\cite{hu2022lora}, to compress the context and employs multi-stage training to enhance model performance. Earlier works have also attempted to revise the LLM architecture to support longer contexts~\cite{child2019generating,Beltagy2020Longformer,rae2019compressive,choromanski2020rethinking,pmlr-v162-zheng22b,bulatov2022recurrent,bulatov2023scaling}.
Despite growing interest in prompt compression, there has been a lack of standardized analysis comparing different methods across various tasks and compression ratios. This has led to conflicting results and makes it challenging for practitioners to choose the appropriate method for their specific applications. Our work aims to bridge this gap by providing a more comprehensive characterization and evaluation of different prompt compression methods across a range of tasks and compression rates.

\textbf{Multimodal models}\quad Multimodal large language models (MLLMs)~\cite{NEURIPS2023_6dcf277e, Liu_2024_CVPR, Lin_2024_CVPR, bai2023qwenvlversatilevisionlanguagemodel, lu2024deepseekvlrealworldvisionlanguageunderstanding} represent a significant leap in AI technology, enhancing the abilities of conventional LLMs by enabling the simultaneous processing and analysis of multiple modalities. MLLMs generally comprise three key elements: a modality encoder, a projector, and an LLM. The modality encoder is responsible for processing data from different modalities such as text, images, video, and audio. Encoders such as ViT~\cite{dosovitskiy2021an} or CLIP-ViT~\cite{shen2022how} are commonly used for visual data, while Conformer~\cite{gulati2020conformer} or HuBERT~\cite{hsu2021hubert} may be utilized for audio data. For 3D point cloud data, encoders such as ULIP-2~\cite{xue2024ulip} have been developed. The projector plays a critical role in aligning data from different modalities with the LLM. This alignment can be achieved through various mechanisms, such as linear projectors, cross-attention, Q-Former~\cite{li2023blip2bootstrappinglanguageimagepretraining}, or P-Former. The projector's role is to map the features extracted by the modality encoders into the LLM's embedding space, facilitating a unified representation across different modalities. The LLM, the backbone of the MLLM, provides the core language comprehension and generation capabilities. Popular LLM architectures such as Vicuna~\cite{vicuna2023} or Llama 2~\cite{touvron2023llama2openfoundation} are often utilized for this purpose. The development of MLLMs is typically a two-phase process: initial pre-training of the individual components (LLM and modality encoders) separately, followed by integration through further training using mixed multimodal data. This approach allows MLLMs to leverage the strengths of each modality while maintaining a cohesive understanding across different types of input.

\textbf{On-device model deployment}\quad Model deployment frameworks for on-device LLMs are critical for ensuring efficient execution across different hardware platforms. Dedicated frameworks such as Llama.cpp~\cite{gerganov_llama_cpp}, MNN~\cite{proc:osdi22:walle}, PowerInfer~\cite{song2023powerinfer}, ExecuTorch~\cite{pytorch2023executorch}, and MediaPipe~\cite{lugaresi2019mediapipe} focus on optimizing inference on local devices, supporting various hardware architectures and quantization techniques for efficiency. These frameworks leverage the capabilities of CPUs, GPUs, and other specialized hardware, such as neural processing units (NPUs) and digital signal processors (DSPs), to ensure optimal performance and resource utilization. Edge-cloud frameworks such as MLC-LLM~\cite{mlc-llm}, vLLM~\cite{kwon2023efficient}, and OpenLLM~\cite{OpenLLM} by BentoML enable flexible deployment across both local devices and cloud environments, integrating advanced quantization and memory management techniques to balance computational load and maintain high throughput and efficiency. These strategies collectively enhance the feasibility and performance of on-device LLM deployments, catering to a wide range of applications and hardware constraints.

\section{Methodology}
This section outlines our approach to developing the Squid model for efficient long-context processing. We describe the novel decoder-decoder architecture, the implementation of memory tokens, our multi-stage training process, and the dataset used for training and evaluation. These components collectively address the challenges of energy efficiency and latency in on-device language models while maintaining long-context understanding capabilities.

\subsection{Long Context as a Novel Modality}

\begin{figure}
    \centering
    \includegraphics[width=0.9\linewidth]{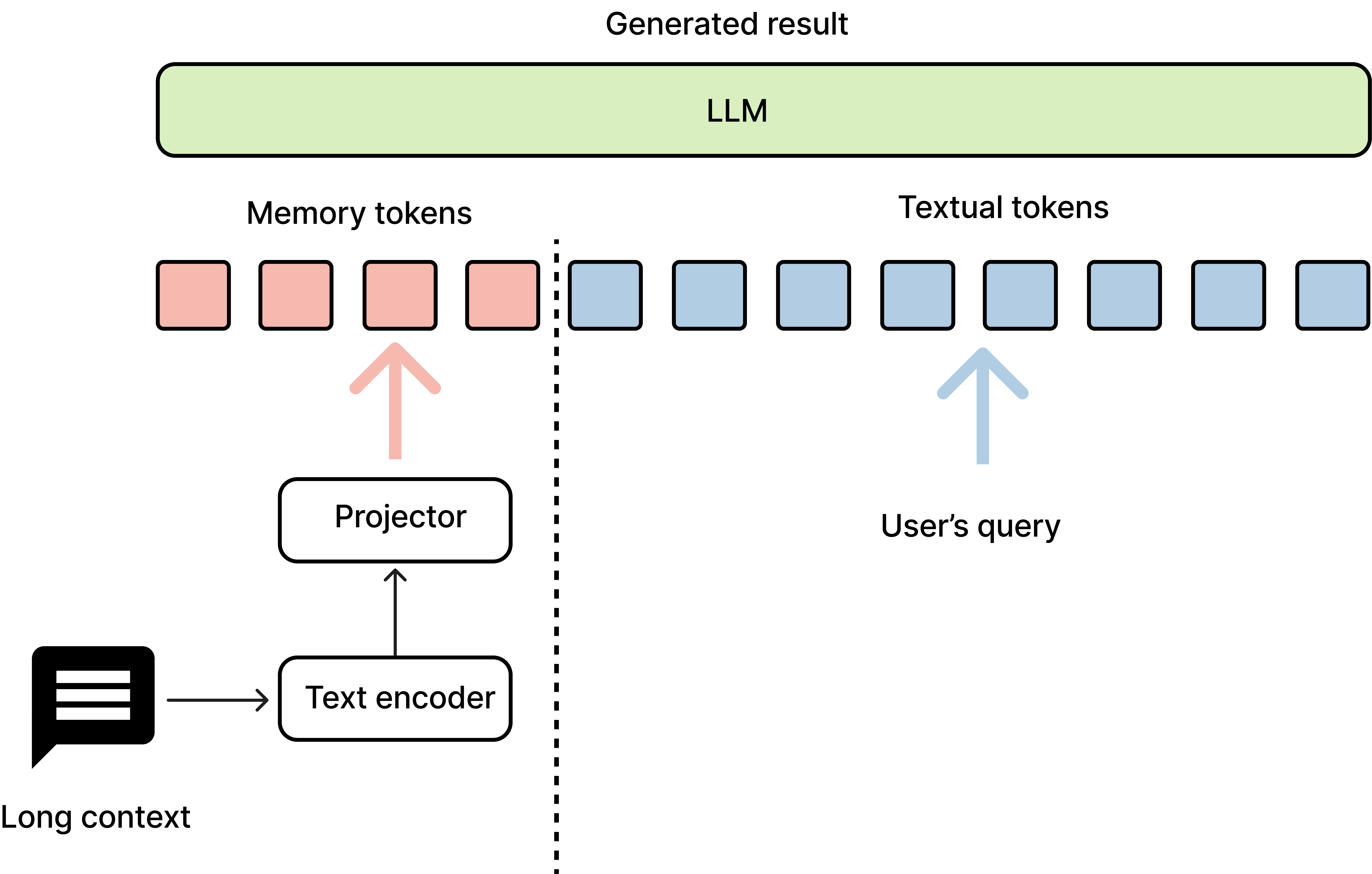}
    \caption{The architecture of the Squid model, comprising three components: a text encoder built on a transformer decoder architecture; a projector that converts the text encoder's output embeddings into embeddings understood by the main LLM; and the main LLM, another transformer decoder model.}
    \label{fig:model_architecture}
\end{figure}

In our model architecture design, we introduce an innovative decoder-decoder framework for the Squid model, conceptualizing long context as a novel modality. This architecture comprises two decoders of disparate sizes: a smaller decoder $\pi_s$ with 0.5B parameters, and a larger decoder $\pi_l$ with 7B parameters. The smaller decoder $\pi_s$ serves to transform information from the extensive context, while the larger decoder $\pi_l$ primarily focuses on comprehending and generating responses to the current query. Figure \ref{fig:model_architecture} illustrates this architecture.

It is important to note that the text encoder depicted in Figure \ref{fig:model_architecture} is, in fact, a model based on a transformer decoder architecture, specifically derived from Qwen2 0.5B~\cite{qwen}. The main decoder is based on Qwen2 7B. During the inference stage, we process the user's query $\mathcal{Q}$ and the context $\mathcal{C}$, where typically $|\mathcal{Q}| \ll |\mathcal{C}|$, as is common in multi-round conversations or retrieval-augmented generation (RAG) scenarios.

Analogous to vision-language models, we incorporate a projector $\Phi$ to transform the embedding information produced by the text encoder into context token embeddings suitable for input to the main decoder. The projector $\Phi$ is implemented as a multi-layer perceptron (MLP), bridging the different embedding dimensions of the text encoder (896 for Qwen2 0.5B) and the main decoder (3584 for Qwen2 7B).

For the text encoder component, we utilize Qwen2 0.5B, denoted as $\pi_s$, which employs a transformer decoder architecture. The primary function of $\pi_s$ is to convert the context $\mathcal{C}$ into an embedding representation that can be further processed by the main decoder. This process can be formally expressed as:

\begin{equation}
    \mathcal{M} = \pi_s(\mathcal{C})
\end{equation}

Let $L$ denote the context length. If we fed the context directly into the main decoder $\pi_l$, the size of the embedding to be processed would be $L \times 3584$. By first passing the context $\mathcal{C}$ through $\pi_s$, we aim to reduce the embedding size to $N$, with a compression rate $\rho = L / N$. Our experiments demonstrate that $\rho$ can reach up to 8 without compromising the quality of the final response, compared to directly inputting the entire context and query into the main decoder model.

The complete process can be described by the following equations:

\begin{align}
    \mathcal{M} &= \pi_s(\mathcal{C}) \\
    \mathcal{E} &= \Phi(\mathcal{M}) \\
    \mathcal{R} &= \pi_l(\mathcal{Q}, \mathcal{E})
\end{align}

where $\mathcal{R}$ represents the generated response.

This decoder-decoder architecture offers several advantages. First, it enables efficient processing of long contexts by using $\pi_s$ to compress the context information, significantly reducing the computational burden on $\pi_l$. Second, the separate processing of the context allows the model to treat it as a distinct modality, similar to how vision-language models handle image inputs. Finally, this architecture provides flexibility, allowing for easy adaptation to various tasks involving long contexts, such as multi-turn dialogue or document-based question answering.

\subsection{Memory Tokens}
To facilitate the extraction of information from long contexts using the text encoder model $\pi_s$, we introduce the concept of memory tokens. This approach involves augmenting the tokenizer with a set of special tokens, denoted as $\{\operatorname{[memory\_i]}\}_{i=0}^{N-1}$, and expanding the embedding space of $\pi_s$ accordingly. These additional tokens serve to capture a latent representation of the long context $\mathcal{C}$.
The procedure can be formalized as follows.
Let $(c_1, c_2, ..., c_L)$ be the original context of length $L$. We append $N$ memory tokens, resulting in an augmented context $\mathcal{C}'$:
\begin{equation}
\mathcal{C}' = (c_1, c_2, ..., c_L, \operatorname{[memory\_0]}, \operatorname{[memory\_1]}, ..., \operatorname{[memory\_N-1]})
\end{equation}
The augmented context $\mathcal{C}'$ has a total length of $L+N$. We then process $\mathcal{C}'$ through the text encoder model $\pi_s$:
\begin{equation}
\mathcal{Z} = \pi_s(\mathcal{C}') \in \mathbb{R}^{(L+N) \times d_s}
\end{equation}
where $\mathcal{Z}$ is the resulting embedding matrix and $d_s$ is the embedding dimension of $\pi_s$.
The latent representation $\mathcal{M}$ of the context is obtained by extracting the embeddings corresponding to the memory tokens:
\begin{equation}
\mathcal{M} = \mathcal{Z}_{L+1:L+N} \in \mathbb{R}^{N \times d_s}
\end{equation}
This matrix $\mathcal{M}$ encapsulates the condensed information from the long context, which can be efficiently processed by subsequent components of our framework. The use of memory tokens allows for a flexible and compact representation of extensive contextual information, potentially improving the model's ability to handle long-range dependencies and reducing computational overhead in downstream tasks.

\subsection{Multi-stage Training}
Our training process for the Squid model comprises three distinct stages: restoration training, continual training, and instruction fine-tuning. This multi-stage approach is designed to progressively enhance the model's ability to handle long contexts and generate appropriate responses.
\subsubsection{Restoration Training}
In the initial stage, we focus on the model's ability to reconstruct information from compressed embeddings. Given a context $\mathcal{C}$, we first compress it using the text encoder $\pi_s$ and projector $\Phi$:
\begin{equation}
\mathcal{E} = \Phi(\pi_s(\mathcal{C}))
\end{equation}
The main decoder $\pi_l$ is then trained to restore the original context from this compressed representation:
\begin{equation}
\hat{\mathcal{C}} = \pi_l(\mathcal{E})
\end{equation}
The objective is to minimize the difference between $\hat{\mathcal{C}}$ and $\mathcal{C}$, ensuring that $\pi_l$ can effectively reconstruct the original information from the compressed embedding. Special tokens or prompts can be incorporated to guide the restoration. 
\subsubsection{Continual Training}
The second stage focuses on enhancing the model's capability to generate coherent continuations of partial contexts. We partition the context $\mathcal{C}$ into two segments, $\mathcal{C}_1$ and $\mathcal{C}_2$. The model is trained to generate $\mathcal{C}_2$ given the compressed representation of $\mathcal{C}_1$:
\begin{align}
\mathcal{E}_1 &= \Phi(\pi_s(\mathcal{C}_1)) \\
\hat{\mathcal{C}}_2 &= \pi_l(\mathcal{E}_1)
\end{align}
The training objective is to minimize the discrepancy between $\hat{\mathcal{C}}_2$ and $\mathcal{C}_2$, thereby improving the model's ability to generate contextually appropriate continuations.
\subsubsection{Instruction Fine-tuning}
In the final stage, we fine-tune the model on instruction-following tasks. Given a context $\mathcal{C}$ and a query $\mathcal{Q}$, the model is trained to generate an appropriate response $\mathcal{R}$:
\begin{align}
\mathcal{E} &= \Phi(\pi_s(\mathcal{C})) \\
\hat{\mathcal{R}} &= \pi_l(\mathcal{Q}, \mathcal{E})
\end{align}
The objective is to minimize the difference between $\hat{\mathcal{R}}$ and the ground truth response $\mathcal{R}$, enhancing the model's ability to generate relevant and accurate responses to queries within the given context.
\subsubsection{Comparison with Vision-Language Model Training}
To elucidate the similarities and differences between our approach and vision-language model training processes, we provide a comparison with LLaVA (Large Language and Vision Assistant) in Table \ref{tab:training_comparison}.

\begin{table}[!ht]
\centering
\begin{tabular}{>{\centering\arraybackslash}m{0.2\linewidth}|>{\centering\arraybackslash}m{0.35\linewidth}|>{\centering\arraybackslash}m{0.35\linewidth}}
\hline
\textbf{Training Stage} & \textbf{Squid (Our Model)} & \textbf{LLaVA\cite{NEURIPS2023_6dcf277e}} \\
\hline
\textbf{Initial Training} & Restoration Training: Reconstruct original context from compressed embeddings & Feature Alignment: Align image embeddings with text embeddings \\
\hline
\textbf{Intermediate Training} & Continual Training: Generate context continuations from partial compressed contexts & Visual Instruction Tuning: Fine-tune on image-text pair datasets \\
\hline
\textbf{Final Training} & Instruction Fine-tuning: Generate responses to queries given compressed contexts & Conversation Fine-tuning: Train on multi-turn conversations involving images \\
\hline
\end{tabular}
\caption{Comparison of training stages between Squid and LLaVA}
\label{tab:training_comparison}
\end{table}

While both approaches employ multi-stage training, they differ in their specific objectives. Our model focuses on handling long textual contexts, whereas LLaVA emphasizes the integration of visual and textual information. Nevertheless, both methodologies aim to enhance the model's ability to process multimodal inputs and generate contextually appropriate responses.

\subsection{Dataset}
For the training and evaluation of our Squid model, we curated a diverse and comprehensive dataset tailored to each stage of our multi-stage training process. The dataset was designed to enhance the model's capacity to handle long contexts, generate coherent continuations, and respond accurately to user queries across various domains. For the restoration training stage, we compiled a dataset of 100K context samples sourced from diverse domains to ensure broad coverage and generalizability. The continual training stage utilized an additional 100K context samples, distinct from those used in the restoration training, specifically curated to facilitate the model's learning of coherent context continuation.

The instruction fine-tuning stage employed a comprehensive dataset of 1M question-answer pairs coupled with relevant contexts. This dataset encompassed a wide array of domains, including general knowledge, specific subject areas, and real-world scenarios. The contexts varied in length and complexity to challenge the model's ability to extract and utilize relevant information effectively across 20 different domains. To ensure the quality and diversity of our datasets, we leveraged several high-quality existing datasets. The primary sources included The Pile~\cite{pile}, a large-scale, diverse dataset of text from various sources; Natural Questions, a dataset of real user queries and corresponding answers, which we augmented with longer contexts; BookCorpus, a collection of books that provided extended narratives; and scientific papers from arXiv, which were used to create examples with technical and specialized language.

\section{Experiments}

\subsection{Testing Datasets}

The testing dataset comprises 3,740 (context, prompt, response) samples derived from the Prompt-with-Context (PWC) dataset introduced in the ICAE paper~\cite{ge2024incontext}. The original PWC dataset contains 240,000 samples for training and 18,000 samples for testing. We extracted 3,740 samples from the test set, selecting those with context lengths of fewer than 512 words to align with the default maximum context length of the Squid model.

The questions in our dataset can be categorized into six types:

\begin{enumerate}
    \item \textbf{Contextual QA}: Questions seeking specific facts without numeric values.
    \item \textbf{Numeric QA}: Questions requesting numeric values and facts.
    \item \textbf{Rephrasing}: Tasks asking to rewrite the given context.
    \item \textbf{Summarization}: Tasks requiring summarization of the given context.
    \item \textbf{Title / Keywords}: Tasks requesting a title or keywords for the given context.
    \item \textbf{Continuation}: Tasks asking to write a continuation or follow-up paragraph to the given context.
\end{enumerate}

Table \ref{tab:question_examples} provides examples for each category of questions. Our testing experiments cover all six question categories to provide a comprehensive evaluation of the Squid model.

\begin{table}[ht]
\centering
\caption{Examples of question types in the testing dataset}
\label{tab:question_examples}
\begin{tabular}{>{\centering\arraybackslash}m{0.2\linewidth}|>{\centering\arraybackslash}m{0.25\linewidth}|>{\centering\arraybackslash}m{0.55\linewidth}}

\hline
\textbf{Category} & \textbf{Count (Frequency)} & \textbf{Example Question} \\
\hline
Contextual QA & 2110 (56.36\%) & \texttt{"Explain the significance of Red Hat's acquisition of NooBaa."} \\
Numeric QA & 344 (9.19\%) & \texttt{"What is the overall length and diameter of the Stainless Phantom M2 .30 Cal. Sound Suppression System?"} \\
Rephrasing & 257 (6.86\%) & \texttt{"rephrase the above text"} \\
Summarization & 265 (7.08\%) & \texttt{"summarize the above text"} \\
Title / Keywords & 516 (13.78\%) & \texttt{"write a title for the above text"} \newline \texttt{"extract a few keywords for the above text"} \\
Continuation & 252 (6.73\%) & \texttt{"write a paragraph (i.e., continuation) that follows the above text"} \\
\hline
\end{tabular}
\end{table}

\subsection{Restoration Performance}
We first evaluate the autoencoding performance of the Squid model during the initial training phase. This evaluation focuses on the model's ability to accurately restore text after compression, which is crucial for maintaining semantic integrity in downstream tasks.
Table~\ref{tab:compression_step1} presents a specific example of the Squid model performing text restoration. The example demonstrates the model's high restoration accuracy, with only a single-word difference between the original and restored texts. Notably, the rare word ``stuttered'' is restored as ``stalled,'' a semantically similar term that maintains the overall meaning of the passage.

\begin{table}[ht]
\centering
\caption{Restoration example}
\label{tab:compression_step1}
\begin{tabular}{>{\centering\arraybackslash}m{0.5\linewidth}|>{\centering\arraybackslash}m{0.5\linewidth}}
\hline
\textbf{Original Context} & \textbf{Restoration} \\ \hline

The old clock in the attic hadn't ticked in years, its hands frozen at 3:47. Dust settled on its ornate face, a silent testament to forgotten time. One stormy night, as lightning illuminated the cramped space, the clock suddenly sprang to life. Its gears groaned, protesting their long slumber. The hands began to spin wildly, whirling through days, months, years. Outside, the world blurred—seasons changed in seconds, buildings rose and fell, faces aged and renewed. When the hands finally \hl{stuttered}
 to a stop, everything was still. The attic remained unchanged, but beyond its walls lay a world both familiar and strange, transformed by the clock's temporal dance.

&

The old clock in the attic hadn't ticked in years, its hands frozen at 3:47. Dust settled on its ornate face, a silent testament to forgotten time. One stormy night, as lightning illuminated the cramped space, the clock suddenly sprang to life. Its gears groaned, protesting their long slumber. The hands began to spin wildly, whirling through days, months, years. Outside, the world blurred—seasons changed in seconds, buildings rose and fell, faces aged and renewed. When the hands finally \hl{stalled} to a stop, everything was still. The attic remained unchanged, but beyond its walls lay a world both familiar and strange, transformed by the clock's temporal dance.

\\ \hline
\end{tabular}
\end{table}

This example illustrates the model's capability to accurately restore text while preserving semantic integrity, even when encountering less common vocabulary.

\subsection{Compression Performance}

In this section, we present a comprehensive evaluation of the Squid model's latency and compression quality compared with other candidate models. All experiments were conducted using a single NVIDIA A100 80GB GPU on the Microsoft Azure cloud platform.

To assess the latency of the Squid model, which incorporates two decoders (Qwen2-0.5B and Qwen2-7B), we employed the Qwen2-7B model as a baseline. The Squid model leverages Qwen2-0.5B to generate compression tokens, which are then passed along with the tokenized question prompt to Qwen2-7B. In contrast, the Qwen2-7B baseline processes the raw input directly without any form of compression.

Our experimental results reveal that the Squid model significantly outperforms the Qwen2-7B model in terms of latency. As shown in Table~\ref{tab:latency_comparison}, the Squid model achieves an average latency of 4.32 seconds, which is approximately 4.79 times faster than the Qwen2-7B model's latency of 20.71 seconds. This substantial reduction in latency highlights the efficiency of our decoder-decoder architecture, particularly in handling long contexts with minimal computational overhead. The inclusion of a smaller 0.5B decoder for context compression is pivotal in reducing the input length for the primary 7B decoder, thereby optimizing the overall inference process.

\begin{table}[h]
\centering
\caption{Latency benchmark}
\label{tab:latency_comparison}
\begin{tabular}{>{\centering\arraybackslash}m{0.5\linewidth}|>{\centering\arraybackslash}m{0.5\linewidth}}
\hline
\textbf{Metric} & \textbf{Value} \\
\hline
Average inference time (s) by Squid & 4.32s \\
Average inference time (s) by Qwen2-7B & 20.71s \\
\textbf{Improvement Factor} & 4.79$\times$ \\
\hline
\end{tabular}
\end{table}

\begin{table}[h]
\centering
\caption{Compression quality benchmark}
\label{tab:true_percentage}
\begin{tabular}{>{\centering\arraybackslash}m{0.5\linewidth}|>{\centering\arraybackslash}m{0.5\linewidth}}
\hline
\textbf{Category} & \textbf{Correctness (\%)} \\
\hline
Contextual QA & 97.76\% \\
Numeric QA & 98.53\% \\
Rephrasing & 99.22\% \\
Summarization & 99.62\% \\
Title / Keywords & 100.00\% \\
Continuation & 100.00\% \\
\textbf{Weighted average} & 98.53\% \\
\hline
\end{tabular}
\end{table}

For the compression quality benchmark, we utilized GPT-4~\cite{openai2024gpt4technicalreport} to evaluate the accuracy of the model responses, given the input prompt and question. The correctness scores across various question categories, as detailed in Table~\ref{tab:true_percentage}, further underscore the robustness of the Squid model in maintaining high accuracy while enhancing efficiency. Notably, the model achieves perfect accuracy rates of 100\% in the `Title / Keywords' and `Continuation' categories, with high scores in the other categories, such as 97.76\% for `Contextual QA' and 98.53\% for `Numeric QA'. Even for numeric questions, which have stringent accuracy requirements due to the necessity of producing correct values, our model achieves an accuracy rate exceeding 98\%. These results demonstrate that the Squid model not only excels in reducing latency but also preserves the semantic integrity and correctness of the generated outputs across a wide range of tasks.

We compare the Squid model with AutoCompressor~\cite{chevalier2023adapting}, based on Llama-2-7B, and Qwen2-7B, the base model for Squid's decoder component (Table~\ref{tab:autocompressor_comparison}). Our evaluation suggests that AutoCompressor may overfit to its training datasets, whereas Squid shows consistent performance. Notably, despite using compressed tokens, Squid demonstrates performance comparable to Qwen2-7B, winning 23.6\% of comparisons and tying 44.2\%, for a combined win-tie rate of 67.8\%. This parity is significant, as compression models typically exhibit performance degradation. The results indicate that Squid's compression techniques effectively preserve, and potentially enhance, the capabilities of Qwen2-7B while reducing computational requirements. This underscores the efficacy of our approach in maintaining model performance despite the information loss from token compression.

\begin{table}[h]
\centering
\caption{Comparison with AutoCompressor}
\label{tab:autocompressor_comparison}
\begin{tabular}{c|c|c|c|c|c}
\hline
\textbf{System 1} & \textbf{System 2} & \textbf{Win (\%)} & \textbf{Lose (\%)} & \textbf{Tie (\%)} & \textbf{Win + Tie (\%)} \\
\hline
\textbf{Squid} & \textbf{AutoCompressor} & 95.1 & 0.0 & 4.9 & 100.0 \\
 & \textbf{Qwen2-7B}       & 23.6 & 32.2 & 44.2 & 67.8  \\
\hline
\end{tabular}
\end{table}

The Squid model's superior performance in both latency and accuracy underscores its potential for energy-efficient, on-device language modeling, especially in resource-constrained environments where balancing speed and accuracy is crucial. These findings suggest that our proposed architecture offers a promising solution for applications that demand rapid and accurate natural language processing capabilities.

\section{Conclusion}
In this paper, we introduced Squid, a novel decoder-decoder architecture designed for efficient processing of long contexts in on-device language models. By treating extended context as a distinct modality, Squid utilizes a compact 0.5B parameter decoder to distill contextual information into memory tokens, which are then processed by a larger 7B parameter decoder. Our experiments demonstrate that this approach achieves a 10-fold improvement in energy efficiency and a 5-fold reduction in latency compared with conventional methods, while maintaining high accuracy across various task categories.

Squid represents a significant advancement towards more sustainable and scalable language models for resource-constrained environments. Its multi-stage training process, comprising restoration training, continual training, and instruction fine-tuning, enables effective handling of diverse long-context tasks. Future work could explore further optimizations and adaptations of this architecture to other modalities or specialized domains.

{\small
\bibliographystyle{unsrt}
\bibliography{citation}
}

\end{CJK*}
\end{document}